\documentclass{article}

\usepackage{PRIMEarxiv}
\usepackage{float}
\usepackage[utf8]{inputenc} 
\usepackage[T1]{fontenc}    
\usepackage{hyperref}       
\usepackage{url}            
\usepackage{booktabs}       
\usepackage{amsfonts}       
\usepackage{nicefrac}       
\usepackage{microtype}      
\usepackage{lipsum}
\usepackage{fancyhdr}       
\usepackage{graphicx}       
\graphicspath{{media/}}     
\usepackage{soul,color}
\usepackage{multirow}
\usepackage{amsmath}
\usepackage{amsbsy}

\newcommand*\Bell{\ensuremath{\boldsymbol\ell}}
\usepackage{bm}
\title{Deep Learning for Energy Time-Series Analysis and Forecasting
} 

\author{
  Maria Tzelepi, Charalampos Symeonidis, Paraskevi Nousi, Efstratios Kakaletsis, Theodoros Manousis,\\ \bf{Pavlos Tosidis, Nikos Nikolaidis and Anastasios Tefas} \\
  Dept. of Informatics \\
  Aristotle University of Thessaloniki \\
  Thessaloniki, Greece\\
  \texttt{\{mtzelepi,charsyme,paranous,ekakalets,tmanousis,ptosidis,nnik,tefas\}@csd.auth.gr} \\
}

\begin{document}
\maketitle

\begin{abstract}
Energy time-series analysis describes the process of analyzing past energy observations and possibly external factors so as to predict the future. Different tasks are involved in the general field of energy time-series analysis and forecasting, with electric load demand forecasting, personalized energy consumption forecasting, as well as renewable energy generation forecasting being among the most common ones. Following the exceptional performance of Deep Learning (DL) in a broad area of vision tasks, DL models have successfully been utilized in time-series forecasting tasks. This paper aims to provide insight into various DL methods geared towards improving the  performance in energy time-series forecasting tasks, with special emphasis in \textit{Greek Energy Market}, and equip the reader with the necessary knowledge to apply these methods in practice. 
\end{abstract}


\section{Introduction}
\begin{figure}
     \centering
     \includegraphics[width=0.5\linewidth]{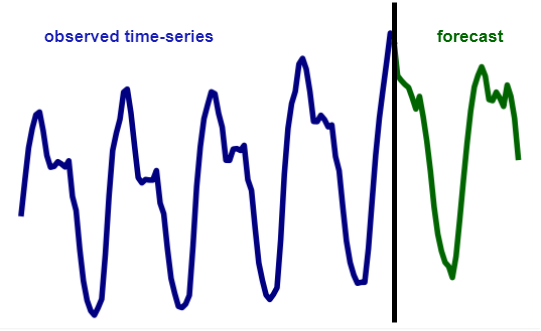}
        \caption{Time-series forecasting.}
        \label{fig:tsf}
\end{figure}

A time-series describes a collection of observations indexed in time order \cite{chatfield2000time}. Time-series forecasting describes the task of predicting future values of a target by analyzing the corresponding past data, as illustrated in Fig. \ref{fig:tsf}. Time-series forecasting involves a broad spectrum of applications, from climate modelling \cite{docheshmeh2022drought} to financial time-series forecasting \cite{fatima2022forecasting} and energy time-series forecasting \cite{alvarez2010energy,torres2022deep,mohammed2022adaptive,hyndman2009density}.

Energy time-series forecasting includes several sub-tasks. In this paper, we focus on three pivotal tasks: \textit{Electric Load Demand Forecasting} (ELDF), describing the process of predicting the electric load demand by employing historical load data,
\textit{Personalized Energy Consumption Forecasting} (PECF), namely the
process of predicting the energy load consumption based on past load data for individual consumers, and \textit{Renewable Energy Generation Forecasting} (REGF), describing the process of predicting mainly the generated solar and wind energy by using, mainly, historical data (e.g., energy generation measurements, weather-related data). 

We deal with the aforementioned forecasting tasks with special emphasis to the \textit{Greek Energy Market}. For each of the tasks, we provide a brief description of earlier important works, as well as more recent Deep Learning (DL) methods, focusing  to recent state-of-the-art methods, considering the aforementioned energy market. That is, considering the ELDF task, we present amongs others, a novel online distillation methodology for improving the forecasting performance of an one-day-ahead DL model, as well as a novel anchored-based methodology. A novel methodology based on the residuals concept  is applied both on ELDF and PECF tasks. Finally, regarding the REGF task a method for wind energy generation prediction guided by multiple-location weather forecasts is presented. In addition, the performance of the presented method is evaluated on other RES types, including solar energy. It should be emphasized that the presented methodologies can also be applied for generic time-series forecasting tasks.

The rest of the manuscript is organized as follows. Section \ref{ELDF} describes the task of Electric Load Demand Forecasting. Subsequently, the task of Personalized Energy Consumption Forecasting is presented in Section \ref{PECF}, while Section \ref{REGF} describes the task of Renewable Energy Generation Forecasting. Finally, Section \ref{conclusions} summarizes this work.

\section{Electric Load Demand Forecasting}
\label{ELDF}
Electric load demand forecasting, also referred to as electricity demand forecasting, describes the process of estimating the amount of electricity that will be used in a certain area, using historical electric load data \cite{singh2013overview}. ELDF constitutes an essential task associated with many critical applications, such as power system operation and planning, or energy trading \cite{jacob2020forecasting}. 

ELDF, based on the time-horizon, falls into three categories: i) short-term, concerning a time-scale of a couple of hours and up to one day or one week ahead, ii) mid-term, concerning a time-scale of one week to one year ahead, and finally iii) long-term, concerning a time-scale of up to a couple of years ahead.

Surveying the relevant literature, we come across various methods ranging from traditional statistical models to modern approaches using machine learning algorithms \cite{singh2013overview}. For example, earlier works include, auto-regressive moving average (ARMA) models \cite{ARMA1, ARMA2}. However, the application of such models comes with some drawbacks, since it prevents the utilization of useful external factors like weather.
Other works include Support Vector Regression (SVR) models \cite{svr_eldf,hong2011svr,setiawan2009very} or random forest-based models \cite{cheng2012random}.

Motivated by the exceptional performance of DL algorithms on a wide range of applications \cite{tzelepi2018deep,tzelepi2021online}, several DL models have been proposed in order to tackle the ELDF task \cite{amarasinghe2017deep,hosein2017load,din2017short,he2017load,lstm1}. For instance,  a convolutional and recurrent neural network based model is proposed for short-term load forecasting in \cite{eskandari2021convolutional}, whilst a recurrent neural network based sequence to sequence model is proposed to forecast the short-term power loads in \cite{zhao2021short}. Furthermore, despite their effectiveness, DL models are often quite sensitive to the used input normalization method, thus, a novel trainable adaptive input normalization method is proposed in \cite{passalis2020global}. The method is able to both maintain important mode information, since global statistics are utilized for the normalization, as well as to consider the current behavior of the time-series and adjust the normalization scheme to this. Experiments performed on two load forecasting datasets validate the effectiveness of the method on such tasks.

Besides, during the recent years several works have been proposed to address the task considering the Greek Energy Market \cite{adamakos2016short,pappas2010electricity,greek3}. For example, a hybrid radial basis function - convolutional neural network model is proposed to address the week-ahead forecasting task in \cite{SIDERATOS_hybrid_2}, while a novel regularization method for improving a lightweight one-day-ahead forecasting model is proposed in \cite{greek1}. Finally, a thorough 
 comparative study of state-of-the-art DL models initially proposed for generic forecasting tasks, applied to the ELDF task on Greek Energy Market is provided in \cite{greek2}

In the following, we briefly present four recent state-of-the-art works on Greek Energy Market. The dataset of Greek Energy Market, used in all the works, consists of historical electric load data and weather data, on an hourly basis, acquired by the Greek Public Power Corporation and OpenWeather\footnote{https://openweathermap.org/}, respectively. Totally seven years of data are utilized. That is, data from years 2012-2016 are used for training, data from 2017 are used for validation, while data from 2018 are used for testing.

\paragraph{Lightweight DL model for Greek Energy Market}
In \cite{greek_lightweight} a simple lightweight model (MLP with two hidden layers) is proposed considering the electricity demand forecasting task considering the Greek energy Market. The authors propose to use as input features the electric load of previous day, week and month, as well as weather information (i.e., temperature values) of these days and temperature forecast for the target day of prediction. Furthermore, three additional values are appended: i) a binary indicator for target day public holiday, ii) a binary indicator for target day being weekend, and iii) an integer value for each day of the week.
The input features are presented in detail in Table \ref{Input Data}. 

Furthermore, in this work, a realistic scenario is developed. More specifically, the vast majority of relevant methods consider that all the historical load data are available. However, in a real-case scenario there is an information gap prior the prediction day. In particular, in Greek Energy Market, there is a gap of 4-10 days. In this work, the authors propose to fill this gap with predictions, before moving on to the final prediction. Finally, a novel loss is proposed targeted at the specific requirements of the Greek Energy Market. 
The authors, provide comparisons of their lightweight model with a state-of-the-art model, i.e., ResNetPlus \cite{chen2018short}, applied on Greek Energy Market. The lightweight model, considering the real-case scenario achieves MAPE 2.52\%, while the significantly more complex ResNetPlus model achieves MAPE 2.77\%.

\begin{table}[h]
	   \caption{Input features for an one-day-ahead model for ELDF}
 	   \label{Input Data}
	    \centering
	  \begin{tabular}{|l|l|l|}
	 \hline    
\textbf{Input Feature} & \textbf{Dim.} &	\textbf{Description} \\ \hline
$\mathbf{l}_{d-1}$ & 24 & Load of the day that is chronologically one day prior to the prediction day\\\hline 
$\mathbf{l}_{d-7}$ & 24 & Load of the day that is chronologically 7 days prior to the prediction day\\ \hline
$\mathbf{l}_{d-28}$ &	24 & Load of the day that is chronologically 28 days prior to prediction day\\\hline
$\mathbf{t}_{d-1}$ & 24 & Corresponding temperature for $\mathbf{l}_{d-1}$\\ \hline
$\mathbf{t}_{d-7}$	& 24	& Corresponding temperature for $\mathbf{l}_{d-7}$\\ \hline
$\mathbf{t}_{d-28}$ & 24 & Corresponding temperature for $\mathbf{l}_{d-28}$\\ \hline
$\mathbf{t}_{d}$ & 24 & Corresponding temperature forecast for prediction day \\\hline
$H$ & 1 & Binary indicator of prediction day being holiday \\ \hline
$W$ & 1 & Binary indicator of prediction day being weekend\\ \hline
$D$ & 1 & Indicator for which day of the week the prediction day is  \\ \hline
    \end{tabular}
	    \end{table}

\paragraph{AFORE: Anchored-based FOREcasting}
Anchored-based FOREcasting (AFORE) \cite{10096754} method draws inspiration by the use of anchors or predefined boxes in object detection methods \cite{ren2015faster,liu2016ssd,he2017mask,lin2017focal}. In these approaches, the model is provided with the aforementioned anchors, and instead of predicting the absolute coordinates of object of interest, the detection task is reformulated as offset prediction with respect to the best matching anchor.  In this way the task is simplified, rendering it easier for the model to learn.

Correspondingly, considering the one-day-ahead electricity demand task, AFORE proposes to define an anchor (i.e., load one week prior the day whose load demand we aim to predict), and subsequently a forecasting model is trained, instead of learning the actual load values, to learn the offset. That is, the learning objective in our task, is reformulated as learning the percentage change of the load with respect to the load one week before.

The anchored targets $\mathbf{\Bell}^{pct}_d$ for the target day $d$ are mathematically formulated as follows: \\
\begin{equation}
    \mathbf{\Bell}^{pct}_d = \frac{\mathbf{\Bell}_d}{\alpha} - 1,
\end{equation}
where $\mathbf{\Bell}_d \in \mathbb{R}^{24}$ refers to the electric load demand of the day whose load demand we aim to predict, $\alpha \in \mathbb{R}^{24}$ refers to the anchor, which is defined as $\alpha=\mathbf{\Bell}_{d-7}$. 

Thus, during the training phase the model learns a percentage change $\mathbf{\hat{\Bell}}^{pct}_d$ by minimizing a loss function. Considering anchored-based methods, usually smooth l1 loss, $\mathcal{L}_{\mbox{s-l1}}$, is used: 
\begin{equation}
    \mathcal{L}_{\mbox{s-l1}}(\mathbf{\Bell}^{pct}_d,\mathbf{\hat{\Bell}}^{pct}_d) = 
    \begin{cases}
      0.5 (\mathbf{\Bell}^{pct}_d-\mathbf{\hat{\Bell}}^{pct}_d)^2 , & \text{if } |\mathbf{\Bell}^{pct}_d-\mathbf{\hat{\Bell}}^{pct}_d| < 1 \\
       |\mathbf{\Bell}^{pct}_d-\mathbf{\hat{\Bell}}^{pct}_d| - 0.5 , & \text{otherwise.}
        \end{cases}
\end{equation}

Finally, during the test phase, the predictions are converted back to the original load space so as to be compared with the ground truth targets of actual load and evaluate the performance of the proposed method using the Mean Absolute Percentage Error (MAPE). That is, the decoded prediction $\mathbf{\hat{\Bell}}_d$ is formulated as:
\begin{equation}
    \mathbf{\hat{\Bell}}_d = \alpha  \cdot  (\mathbf{\hat{\Bell}}^{pct}_d +1).
\end{equation}
Then, MAPE for a set of $n$ test samples is computed as:
\begin{equation}\label{eq:MAPE}
    \mbox{MAPE} = \frac{100\%}{n} \sum_{d=1}^{n}\bigg|\frac{\mathbf{\Bell}_d - \mathbf{\hat{\Bell}}_d}{\mathbf{\Bell}_d}\bigg|,
\end{equation}
where $\mathbf{\Bell}_d$ is the actual load, while $\mathbf{\hat{\Bell}}_d$ is the model's decoded prediction for the sample $d$.

\begin{figure}[H]
     \centering
     \includegraphics[width=\linewidth]{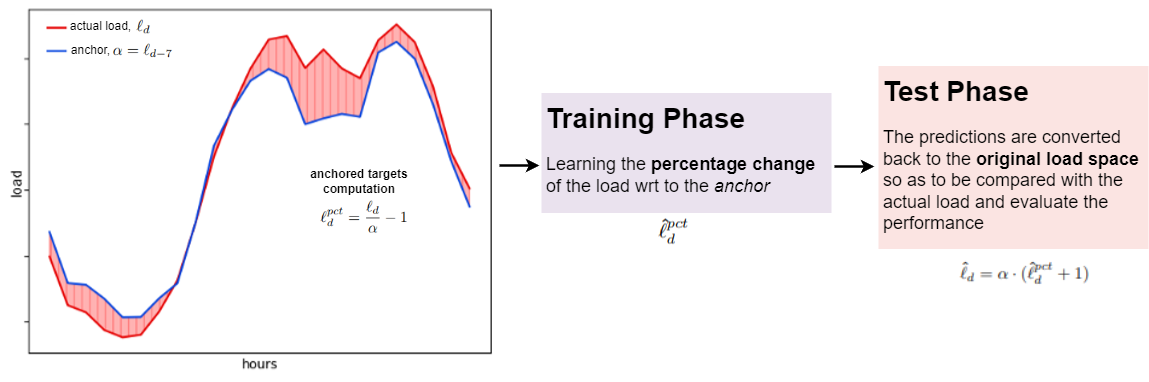}
        \caption{AFORE method: Training and test processes.}
        \label{fig:afore}
\end{figure}

The training and test processes of the AFORE methodology are also illustrated in Fig. \ref{fig:afore}.

Indicative evaluation results validating the efficiency of the AFORE method, considering the one-day-ahead ELDF task (on an hourly basis) on the Greek Energy Market are presented in Tables \ref{table:AFORE-absolute}-\ref{table:AFORE-absolute-achs2}. More specifically, the authors validate the effectiveness of the AFORE method using different loss functions and model architectures respectively.

\begin{table}[!h]
\centering
\caption{AFORE: Test MAPE (\%) for various losses, using a simple MLP, considering the typical setup.} 
\label{table:AFORE-absolute}
\begin{tabular}{|c|c|c|c|}
\hline
\textbf{Method} & \textbf{Smooth L1} & \textbf{L1} & \textbf{L2}  \\ \hline
Baseline & 3.559 $\pm$ 0.050  & 3.557 $\pm$ 0.0863 & 4.034 $\pm$ 0.117 \\ \hline
AFORE & \bf{2.561 $\pm$ 0.104} & \bf{2.782 $\pm$ 0.199} & \underline{\bf{2.518 $\pm$ 0.174}}\\ \hline            
\end{tabular}
\end{table}

\begin{table}[!h]
\centering
\caption{AFORE: Test MAPE (\%) for different model architectures using L2 loss, considering the typical setup.} 
\label{table:AFORE-absolute-achs2}
\begin{tabular}{|c|c|c|}
\hline
\textbf{Method} & \textbf{MLP(2 layers, 64 neurons)} & \textbf{ResNet(2 layers, 64 neurons)}  \\ \hline
Baseline & 2.989 $\pm$ 0.104 &  2.635 $\pm$ 0.056  \\ \hline
AFORE & \bf{2.457 $\pm$ 0.063} &  \underline{\bf{2.387 $\pm$ 0.750 }} \\ \hline        
\end{tabular}
\end{table}

\begin{figure}[H]
     \centering
     \includegraphics[width=\linewidth]{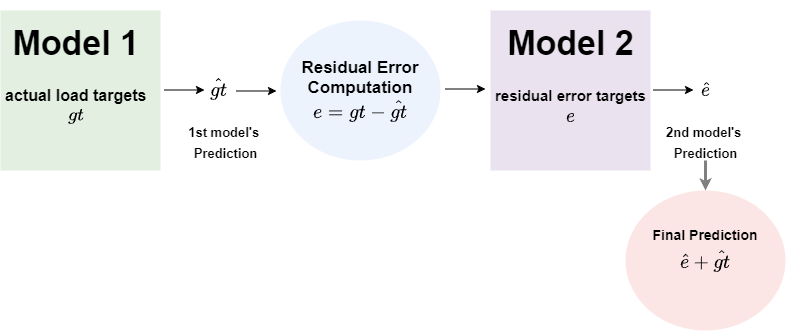}
        \caption{Pipeline of the RESELF methodology.}
        \label{fig:reself}
\end{figure}

\paragraph{RESidual Error Learning for Forecasting}
RESidual Error Learning for Forecasting (RESELF) \cite{tzelepieann} method aims to improve the forecasting performance towards energy time-series forecasting tasks. RESELF utilizes the concept of \textit{residuals}, and proposes the following methodology, as also illustrated in Fig. \ref{fig:reself}. First, a model is trained to predict the actual load values (ground truth), considering the electric load demand forecasting task. Then, the residual errors are computed from the actual load values. In the second stage, the computed residual errors are utilized as targets in order to train a second model. That is, the second model is trained to forecast the residual errors. Finally, the prediction of the RESELF method is formulated as the sum of the first model's and model's predictions. The gist of the method is that if the errors are systematic, then the methodology will provide improved performance. 

Considering the ELDF task on a typical setup where all the past load data and weather information are available, the efficiency of the RESELF method is demonstrated in Table \ref{reself-tab1} in three electricity demand datasets. A simple MLP model with three hidden layers is used in both phases of the RESELF methodology. It should also be highlighted that different models can be utilized in the two phases of the method. More specifically, as authors report, further improvements can be achieved using a relatively more lightweight second model in the RESELF methodology. For example, considering the Greek Energy Market, using a more lightweight second, RESELF method achieves MAPE 2.52\% (against 2.63\% using identical models).

\begin{table}
\centering
\caption{RESELF: Test MAPE (\%) for the RESELF method against baseline.}\label{reself-tab1}
\begin{tabular}{|l|l|l|l|}
\hline
\textbf{Method} &  \textbf{Greek Energy Market} & \textbf{Spain \footnote{https://www.kaggle.com/datasets/nicholasjhana/energy-consumption-generation-prices-and-weather}} & \textbf{ISO-NE \footnote{https://github.com/yalickj/load-forecasting-resnet}}\\
\hline
Baseline & 3.36 $\pm$ 0.08  &  5.62 $\pm$ 0.07 & 2.56 $\pm$ 0.15 \\ \hline
RESELF  & \bf{2.63 $\pm$ 0.15}  &  \bf{4.66 $\pm$ 0.14} &  \bf{2.15 $\pm$ 0.06}\\ 
\hline
\end{tabular}
\end{table}

\paragraph{Online Self-Distillation for Forecasting}
Knowledge Distillation (KD) \cite{hinton2015distilling}  has been proposed as an auspicious methodology for training compact and effective networks by transferring the knowledge obtained usually from more powerful and heavyweight networks. Even though KD has been proven effective, it is associated with some flaws linked with the complex training pipeline, since it requires to train first the so-called teacher network, and after its convergence to transfer the obtained knowledge to the so-called student network. Hence, more recent works propose a single-stage pipeline by omitting the teacher's pre-training process. This methodology is known as online distillation.

Event though (online) distillation has been extensively studied with respect to classification tasks, the research considering forecasting tasks is extremely limited, despite the fact that the underlying reasons entailing KD, associated with requirement for compact and fast networks, are also apparent in these tasks. In a recent work \cite{9976318}, the authors introduce the concept of online self distillation to the ELDF task. More specifically, they exploit the idea that it is advantageous to train a network with the so-called soft targets, as opposed to training it only with the ground truth targets (hard targets). 

Correspondingly, considering the electricity demand forecasting task, it is advocated that, since similar input features lead to similar predicted load values, the forecasting ability of the network can be improved by incorporating these similarities to the training process, as opposed to training with the actual load values. 

This can be mathematically articulated as follows. Considering a neural network for electricity demand forecasting, $\phi(\cdot\, ; \mathcal{W})$ with weights $\mathcal{W}$, an input sample $\mathbf{x}_i, \; i=1, \cdots, N,$ its corresponding output of the network, $ \phi(\mathbf{x}_i, \mathcal{W})$, and its ground truth vector $\mathbf{g}_i\in \mathbb{R}^{d}$, the soft target $\mathbf{s}_i$ for the input sample in the distilled training process is computed as a combination of the ground truth vector and the prediction of the model as follows:
\begin{equation}\label{eq:soft_target}
   \mathbf{s}_i = \mathbf{g}_i + \lambda  \phi(\mathbf{x}_i, \mathcal{W}), 
\end{equation}
where $\lambda \in (0,1)$ is used to control the contribution of the two components.

Therefore, the authors propose to train the forecasting model with the soft target $\mathbf{s}_i \in \mathbb{R}^{d}$, instead of the ground truth target $\mathbf{g}_i$. That is, using MAPE loss, a common loss considering forecasting tasks for training the model, the loss $\mathcal{L}_{osdf}$ of method is formulated as follows, utilizing the computed soft targets: 
\begin{equation}\label{eq:osdf_loss}
    \mathcal{L}_{osdf} = \frac{1}{N}\sum_{i=1}^N \bigg|\frac{ \mathbf{g}_i -(1-\lambda)  \phi(\mathbf{x}_i, \mathcal{W})}{ \mathbf{g}_i + \lambda  \phi(\mathbf{x}_i, \mathcal{W})}\bigg|.
\end{equation}

Considering the one-day-ahead ELDF task on Greek Energy Market, the OSDF method improves the baseline performance of a simple MLP model using the input features that reported in \cite{greek_lightweight}, in terms of MAPE. For example, considering a realistic setup where there is a gap in past load data, the baseline method of training without distillation achieves MAPE 5.967\%, while the OSDF method achieves MAPE 5.754\%.  
 Finally, it should be noted that the aforementioned online distillation method can be extended so as to acquire the additional knowledge from the model itself using the neighborhood of each sample in the output of the forecasting model. That is, the soft targets can be computed as the actual load values and the predictions of the most similar, in terms of a similarity metric, samples. 

\section{Personalized Energy Consumption Forecasting}\label{PECF}
A similar to the ELDF task is the personalized energy consumption forecasting, which aims at improving the individuals' energy consumption behavior.  PECF can concern low, medium, and high voltage consumers. Several works have been proposed so as to address the PECF task during the recent years
\cite{mckerracher2013energy,wang2022personalized,7366172,KHAN2021107023,LEI2021110886}. 
More recently, advances in energy technologies allowed consumers to manage energy consumption in their homes. The so-called Home Energy Management (HEM) constitutes an essential task during the recent few years, with a growing research interest \cite{gomes2022recent}.

In the following, we present a recent work considering the Greek Energy Market \cite{tzelepieann}. More specifically, simple lightweight models have been proposed to address the energy consumption forecasting for Medium Voltage (MV), and High Voltage (HV) consumers, considering one short-term and one mid-term forecasting tasks.

The \textit{MV/HV Personalized Consumption} dataset that used in this work consists of historical load data, provided by the Greek Public Power Corporation, and is based on MV and HV individual customer load data, as well as weather information, obtained from OpenWeather.

Two specific forecasting tasks are addressed in \cite{tzelepieann}, that is one-day-ahead and one-year-ahead forecasting tasks. Simple MLP models are utilized consisting of three hidden layers. The input features for one-day-ahead task have been presented in Table \ref{Input Data}. Considering the one-year-ahead task, the input features are presented in Table \ref{table:Personalized_Input_Data-year}. Note also, that following a similar logic, an one-month-ahead forecasting model can be developed. The input features are presented in Table \ref{table:Personalized_Input_Data-month}.

\begin{table}[!ht] 
	   \caption{Description of input features - One-year-ahead forecasting task.}
 	   \label{table:Personalized_Input_Data-year}
	    \centering
	  \begin{tabular}{|l|l|l|}
	       \hline
\textbf{Input Feature} & \textbf{Dim.} & \textbf{Description} \\ \hline
$L_{m}$ & 12 & Consumption of previous 12 months\\\hline
$T_{m}$ & 12 & Temperature of previous 12 months\\ \hline
$Hu_{m}$ & 12 & Humidity of previous 12 months\\ \hline
$S_{m}$ & 12 & Season\\ \hline
$H_{m}$ & 12 & Holiday\\ \hline
$M_{m}$ & 12 & Month\\ \hline
$Y_{m}$ & 12 & Year\\ \hline
Avg & 1 & Average Consumption of previous 12 months\\ \hline
Std & 1 & Standard Deviation of Consumption previous 12 months\\ \hline
Min & 1 & Minimum consumption of previous 12 months\\ \hline
Max & 1 & Maximum consumption of previous 12 months\\ \hline
Skew & 1 & Skewness of consumption of previous 12 months\\ \hline
 	  \end{tabular}
\end{table}

\begin{table}[!ht] 
	   \caption{Description of input features - One-month-ahead forecasting task.}
 	   \label{table:Personalized_Input_Data-month}
	    \centering
	  \begin{tabular}{|l|l|l|}
	       \hline
\textbf{Input Feature} & \textbf{Dim.} & \textbf{Description} \\ \hline
$L_{d}$ & 10 & Consumption of previous 10 days \\ \hline
$T_{d}$ & 10 & Temperature of previous 10 days  \\ \hline
$Hu_{d}$ & 10 & Humidity  of previous 10 days \\ \hline
$S_{d}$ & 10 & Season\\ \hline
$Y_{d}$	& 10 & Year \\ \hline
$W_{d}$ & 10 &  Week\\ \hline
$DW$ & 10 & Day of week \\ \hline
$DY$ & 10 & Day of year \\ \hline
$DM$ & 10 & Day of month \\ \hline
$H$ & 10 & Holiday\\\hline
$M$ & 10 & Month \\ \hline
Avg & 1 & Average Consumption of previous 10 days\\ \hline
Std & 1 & Standard Deviation of Consumption of previous 10 days\\ \hline
Min & 1 & Minimum consumption of previous 10 days\\ \hline
Max & 1 & Maximum consumption of previous 10 days\\ \hline
Skew & 1 & Skewness of consumption of previous 10 days\\ \hline
 	  \end{tabular}
	    \end{table}

Finally, in \cite{tzelepieann} the RESELF method for addressing the ELDF task, has also been applied for addressing the PECF task. The evaluation results, in terms of MAPE, considering one MV consumer and one HV consumer are presented in Table \ref{reself-tab2}, validating the efficiency of the RESELF methodology.

\begin{table}
\centering
\caption{RESELF: Test MAPE (\%) for the RESELF method against baseline for two MV and HV consumers, considering the one-year-ahead prediction task.}\label{reself-tab2}
\begin{tabular}{|l|l|l|l|}
\hline
\textbf{Method} & \textbf{MV} & \textbf{HV}\\
\hline
Baseline & 8.27   &  1.66 \\ \hline
RESELF  & \bf{6.19}  &  \bf{1.60}\\
\hline
\end{tabular}
\end{table}

\section{Renewable Energy Generation Forecasting}\label{REGF}
Renewable energy generation forecasting, also referred to as Renewable Energy Sources (RES) forecasting,  concerns mainly the prediction of the generated solar and wind energy, within a specific time frame. The accurate prediction of those two dominant RES, is a vital component of the management and operation of electric grids.
Many techniques for solar energy prediction of photovoltaic facilities using historical data are presented in the recent literature  ~\cite{1_ahmed2020review,2_gandhi2020review,3_pierro2021machine}. Moreover, several works present long and short-term wind energy prediction approaches via recurrent neural networks models~\cite{4_woo2018predicting, 5_he2014spatio, 6_barbounis2006long, 7_noorollahi2016using}, convolutional neural networks (CNNs)~\cite{8_yu2019scene, 9_zhu2020_cnn} and autoencoders~\cite{10_wang2021autoenc}.

More recently, attention-based models have also gathered a lot of interest in REGF. In~\cite{11_niu2020att}, the authors proposed a sequence-to-sequence model for multi-step-ahead wind power forecasting. The model architecture consists of two groups of Gated Recurrent Unit (GRU) blocks, which work as encoder and decoder.  The authors proposed the Attention-based GRU (AGRU) for embedding the task of correlating different forecasting steps by hidden activations of GRU blocks. In addition, the authors in~\cite{12_santos2022tft} applied the Temporal Fusion Transformer (TFT)~\cite{13_lin2021tft}, an attention-based architecture, aiming to predict the hourly day-ahead phtotovoltaic (PV) power generation. TFT was able to learn long-term and short-term temporal relationships respectively, and also build efficiently feature representations of static variables, observed and known time-varying inputs. These components allowed the model to outperform state-of-the-art methods like LSTMs.

As far as the Greek Energy market is concerned, the authors in ~\cite{15_vartholomaios2021short} present Short-term Renewable Energy Generation Forecasting in Greece using Prophet Decomposition and Tree-based Ensembles. More precisely, this paper proposes a new dataset for solar and wind energy generation forecast in Greece and introduces a feature engineering pipeline that enriches the dimensional space of the dataset. In addition, it proposes a novel method that utilizes the innovative Prophet model, an end-to-end forecasting tool that considers several kinds of nonlinear trends in decomposing the energy time series before a tree-based ensemble provides short-term predictions. The proposed method manages to outperform all competing baseline methods, presenting
both lower error rates and more favorable error distribution.

Reviewing further the recent literature, the authors in ~\cite{16_tosounidis2022wind} evaluated the performance of a series of deep learning models on the task of wind energy forecasting using a varying number of horizon time-steps. For this task, the authors utilized Renewable Energy Sources (RES) data collected from wind-turbines in the Greek and Portuguese regions, containing power production values, with a frequency of 10 minutes, as well as on-site weather-related measurements. In similar fashion, the authors in ~\cite{17_papaioannou2022renewable}, evaluated the performance of forecasting method on solar energy forecasting on 1-hour and 24-hours prediction windows. The data, employed in the corresponding experimental evaluation, were collected from a single solar park in Greece. Additionally, the authors conducted an analysis on the effect of the parameters of the employed DL-models, on their forecasting performance.

Finally, the authors in ~\cite{18_petrosrenewable} also reviewed several machine learning and deep learning algorithms for REGF.  This particular case study makes use of two datasets, one for solar energy forecasting and another for wind energy forecasting. Both datasets span between years 2015-2017 and contain the actual solar and wind energy production in Greece. The corresponding forecasting methodologies were separated into three main categories according to the type of prediction: the short-term prediction of the generation of the next hour, the prediction of a predetermined forecast horizon in a multi-output approach and the prediction of the generation several hours ahead in a recursive manner. These methodologies and features, that can be used in each approach to best represent the determining factors of the REGF, are applied in a plethora of machine learning algorithms, such as Random Forest Regressor, Gradient Boosting Regressor, XGBoost Regressor and Support Vector Regressors, as well as many deep learning networks, using Long Short-Term Memory networks and Convolutional Neural Networks with many evaluation metrics.

In the following, we briefly present a recently proposed state-of-the-art method for the REGF task on RES data collected from Greece.

\begin{figure}
     \centering
     \includegraphics[width=0.5\linewidth]{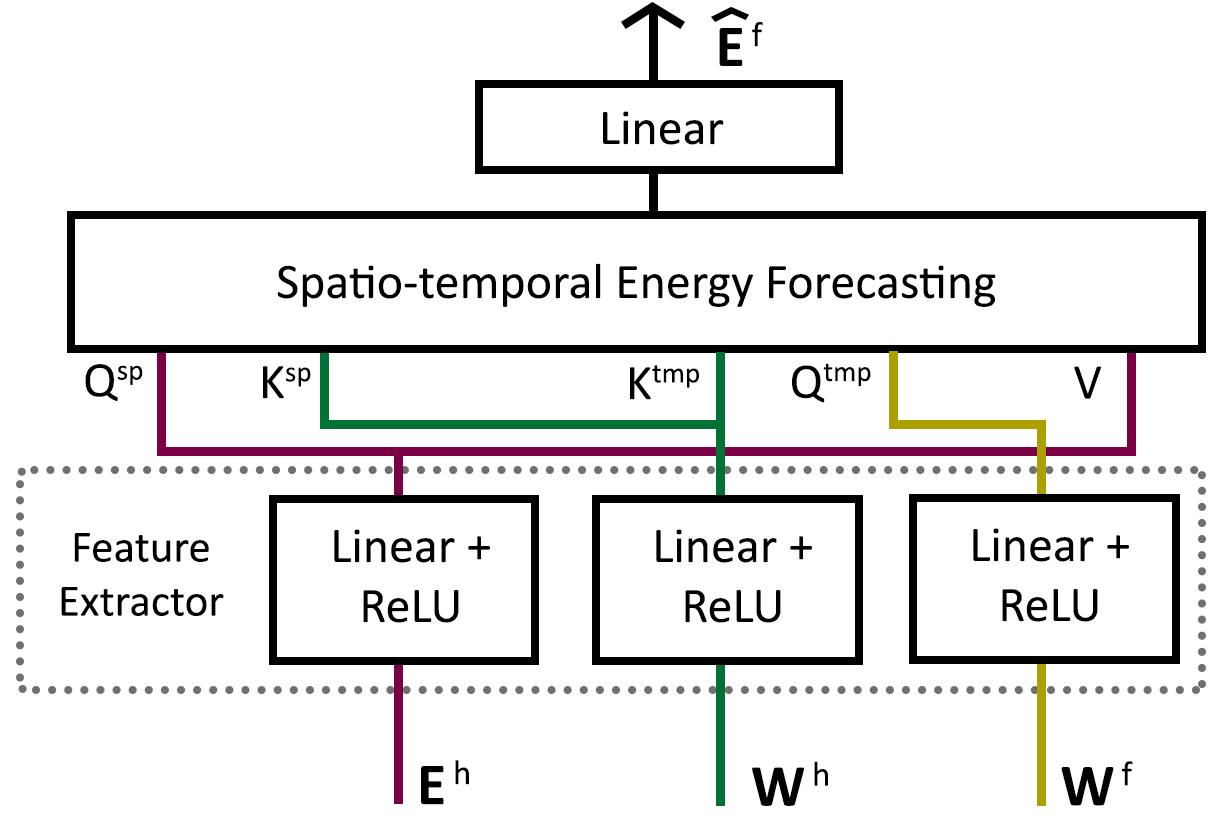}
        \caption{An abstract illustration of the pipeline employed in \cite{14_symeonidis2023wind}, proposed for wind energy generation forecasting.}
        \label{fig:res_for_arch}
\end{figure}

\paragraph{Wind Energy Generation Prediction Guided by Multiple-Location Weather Forecasts}  In ~\cite{14_symeonidis2023wind} the authors presented a method suitable for short-term wind energy forecasting guided by multiple-location weather forecasts. The presented method, relies on a novel variant of the scaled-dot product attention mechanism, for exploring relations between the generated energy and a set of multiple-location weather forecasts/measurements. An illustration of the pipeline of the method is depicted in Figure ~\ref{fig:res_for_arch}. The conducted experimental evaluation on a dataset consisting of the hourly generated wind energy in Greece along with hourly weather estimations for 18 different locations outperformed several SoA time-series forecasting methods.

Aiming to explore the efficacy of REGF methods, four different datasets, from various RES types in Greece were formalized. Details about those datasets are provided in Table~\ref{Tab:res_datasets}. More specifically, datasets D.1 and D.2  consist of High Voltage (HV) wind and solar energy generation measurements from Greece, at hour-level resolution, respectively. Those  measurements have been retrieved from ENTSO-E~\footnote{\url{https://transparency.entsoe.eu/dashboard/show}}. Overall, those two datasets span between 01.01.2017 and 31.01.2023. The rest of the formalized datasets, namely D.3 and D.4, consist of energy measurements, at hour-level resolution, from renewable energy sources in Greece, without an explicit separation of the type of the RES. More specifically, D.3 contains HV energy generation measurements, while D.4 contains Medium Voltage (MV) and Low Voltage (LV) energy generation measurements. The energy generation measurements have been retrieved by Independent Power Transmission Operator (IPTO)~\footnote{\url{https://www.admie.gr/agora/statistika-agoras/dedomena}}. The largest percentage of D.3 is generated from wind and solar power stations in Greece. Additionally, the largest percentage of D.4 is generated from Photovoltaic (PV) panels installed by the users within the power grid. Overall, datasets D.3 and D.4  span between 01.11.2020 and 8.01.2023. 

\begin{table}[!h]
\caption{Four datasets suitable for REGF in Greece.}
\label{Tab:res_datasets}
\resizebox{\columnwidth}{!}{
\begin{tabular}{|l|l|l|l|l|}
\hline
 \textbf{Dataset ID} & \textbf{Type of RES} & \textbf{Source}  & \textbf{Time Period of Training Set} & \textbf{Time Period of Testing Set} \\ \hline
 D.1 & Wind (HV) & ENTSO-E & 01.01.2017 - 31.12.2021 & 01.01.2022 - 31.01.2023 \\ \hline
 D.2 & Solar (HV) & ENTSO-E & 01.01.2017 - 31.12.2021 & 01.01.2022 - 31.01.2023 \\ \hline
 D.3 & RES TSO System (HV) & IPTO & 01.11.2020 - 31.05.2022 & 1.06.2022 - 08.01.2023 \\ \hline
 D.4 & RES TSO Network (MV \& LV) & IPTO & 01.11.2020 - 31.05.2022 & 1.06.2022 - 08.01.2023  \\ \hline
\end{tabular}
}
\end{table}

All employed datasets are accompanied by weather data estimations from the OpenWeather\footnote{\url{https://openweathermap.org/}} online service. The corresponding data include estimations of various weather measurements and conditions, such as wind speed and direction, temperature, humidity and cloudiness, from 29 locations in Greece. Those locations coincide with locations of large solar and wind power stations in Greece.    

Two experimental REGF scenarios were formalized. The first scenario simulates a day-ahead REGF scenario, while the second simulates a week-ahead REGF scenario. In both scenarios, a 48-hour gap between the input window and the forecasting window is introduced, aiming to simulate the case where the database is not up-to-date with the energy measurements of the last two days. Details about the REGF scenarios are provided in Table~\ref{Tab:res_scenarios}.

\begin{table}[!h]
\caption{Day-ahead and week-ahead REGF scenarios.}
\label{Tab:res_scenarios}
\resizebox{\columnwidth}{!}{
\begin{tabular}{|l|l|l|l|}
\hline
\textbf{Experimental Evaluation} & \textbf{Input Window} & \textbf{Prediction Window}  & \textbf{Duration Between Input Window} \\ 
\textbf{Scenario} & \textbf{Size (Hours)} & \textbf{Size (Hours)} & \textbf{and Prediction Window (Hours)}\\
 \hline
 Day-ahead energy forecasting & 72 & 24 & 48 \\ \hline
 Week-ahead energy forecasting & 336 & 168 & 48 \\ \hline
\end{tabular}}
\end{table}

 TFT~\cite{13_lin2021tft} and ~\cite{14_symeonidis2023wind} were employed for REGF, in the experimental evaluation process. Though ~\cite{14_symeonidis2023wind} was employed, by the corresponding authors, specifically for wind energy generation forecasting, we evaluate the method's performance on the general task of REGF without making any significant modification in its architecture and pipeline. Mean Absolute Error (MAE) and Root Mean Square Error (RMSE) were selected as evaluation metrics. The performance of the two employed methods on datasets D.1-D.4 are presented on Tables~\ref{Tab:res_d1},~\ref{Tab:res_d2},~\ref{Tab:res_d3} and ~\ref{Tab:res_d4}, respectively. In almost all experimental setups, ~\cite{14_symeonidis2023wind} was able to achieve top results, compared to ~\cite{13_lin2021tft}, demonstrating the effectiveness of exploring spatio-temporal relationships between the energy-related and weather-related data.   

\begin{table}[!h]
\centering
\caption{Test MAE and RMSE of REGF methods in D.1 dataset.}
\label{Tab:res_d1}
\resizebox{\columnwidth}{!}{
\begin{tabular}{|l|ll|ll|}
\hline
\multirow{2}{*}{\textbf{Method}} & \multicolumn{2}{l|}{\textbf{Day-ahead energy forecasting}} & \multicolumn{2}{l|}{\textbf{ Week-ahead energy forecasting }}  \\ \cline{2-5} 
& \multicolumn{1}{l|}{\textbf{MAE (nrm. / MW)}} & \textbf{RMSE (nrm. / MW)} & \multicolumn{1}{l|}{\textbf{MAE (nrm. / MW) }} & \multicolumn{1}{l|}{\textbf{RMSE (nrm. / MW)}} \\ \hline
\cite{13_lin2021tft} & \multicolumn{1}{l|}{0.102 / 309.88} &  0.135 / 413.07  & \multicolumn{1}{l|}{0.113 / 344.99 } & \multicolumn{1}{l|}{0.140 / 428.34 } \\ \hline
\cite{14_symeonidis2023wind} & \multicolumn{1}{l|}{\textbf{0.094} / \textbf{285.76} } & \textbf{0.126} / \textbf{383.46} & \multicolumn{1}{l|}{\textbf{0.090} / \textbf{275.99}} & \multicolumn{1}{l|}{\textbf{0.120} / \textbf{367.58}} \\ \hline
\end{tabular}}
\end{table}

\begin{table}[!h]
\centering
\caption{Test MAE and RMSE of REGF methods in D.2 dataset.}
\label{Tab:res_d2}
\resizebox{\columnwidth}{!}{
\begin{tabular}{|l|ll|ll|}
\hline
\multirow{2}{*}{\textbf{Method}} & \multicolumn{2}{l|}{\textbf{Day-ahead energy forecasting}} & \multicolumn{2}{l|}{\textbf{ Week-ahead energy forecasting }}  \\ \cline{2-5} 
& \multicolumn{1}{l|}{\textbf{MAE (nrm. / MW)}} & \textbf{RMSE (nrm. / MW)} & \multicolumn{1}{l|}{\textbf{MAE (nrm. / MW) }} & \multicolumn{1}{l|}{\textbf{RMSE (nrm. / MW)}} \\ \hline
\cite{13_lin2021tft} & \multicolumn{1}{l|}{ 0.075 / 356.60} &  0.137 / 647.76  & \multicolumn{1}{l|}{ 0.074 / 352.81 } & \multicolumn{1}{l|}{ 0.134 / 633.06} \\ \hline
\cite{14_symeonidis2023wind} & \multicolumn{1}{l|}{\textbf{0.024} / \textbf{114.28}} & \textbf{0.049} / \textbf{229.99} & \multicolumn{1}{l|}{\textbf{0.023} / \textbf{108.59}} & \multicolumn{1}{l|}{ \textbf{0.046} / \textbf{218.13}} \\ \hline
\end{tabular}}
\end{table}

\begin{table}[!h]
\centering
\caption{Test MAE and RMSE of REGF methods in D.3 dataset.}
\label{Tab:res_d3}
\resizebox{\columnwidth}{!}{
\begin{tabular}{|l|ll|ll|}
\hline
\multirow{2}{*}{\textbf{Method}} & \multicolumn{2}{l|}{\textbf{Day-ahead energy forecasting}} & \multicolumn{2}{l|}{\textbf{ Week-ahead energy forecasting }}  \\ \cline{2-5} 
& \multicolumn{1}{l|}{\textbf{MAE (nrm. / MW)}} & \textbf{RMSE (nrm. / MW)} & \multicolumn{1}{l|}{\textbf{MAE (nrm. / MW) }} & \multicolumn{1}{l|}{\textbf{RMSE (nrm. / MW)}} \\ \hline
\cite{13_lin2021tft} & \multicolumn{1}{l|}{ 0.100 / 336.90} & \textbf{0.117} / \textbf{392.33}  & \multicolumn{1}{l|}{ 0.100 / 338.76 } & \multicolumn{1}{l|}{ 0.135 / 446.51} \\ \hline
\cite{14_symeonidis2023wind} & \multicolumn{1}{l|}{\textbf{0.094} / \textbf{318.95}} & 0.120 / 399.14 & \multicolumn{1}{l|}{ \textbf{0.092} / \textbf{313.37}} & \multicolumn{1}{l|}{\textbf{0.122} / \textbf{405.33}} \\ \hline
\end{tabular}}
\end{table}

\begin{table}[!h]
\centering
\caption{Test MAE and RMSE of REGF methods in D.4 dataset.}
\label{Tab:res_d4}
\resizebox{\columnwidth}{!}{
\begin{tabular}{|l|ll|ll|}
\hline
\multirow{2}{*}{\textbf{Method}} & \multicolumn{2}{l|}{\textbf{Day-ahead energy forecasting}} & \multicolumn{2}{l|}{\textbf{ Week-ahead energy forecasting }}  \\ \cline{2-5} 
& \multicolumn{1}{l|}{\textbf{MAE (nrm. / MW)}} & \textbf{RMSE (nrm. / MW)} & \multicolumn{1}{l|}{\textbf{MAE (nrm. / MW) }} & \multicolumn{1}{l|}{\textbf{RMSE (nrm. / MW)}} \\ \hline
\cite{13_lin2021tft} & \multicolumn{1}{l|}{ 0.078 / 354.14} & 0.126 / 517.05 & \multicolumn{1}{l|}{ 0.064 / 309.18} & \multicolumn{1}{l|}{ 0.096 / 416.70} \\ \hline
\cite{14_symeonidis2023wind}  & \multicolumn{1}{l|}{\textbf{0.037} / \textbf{216.58}} & \textbf{0.053} / \textbf{271.00} & \multicolumn{1}{l|}{\textbf{0.039} / \textbf{224.35}} & \multicolumn{1}{l|}{\textbf{0.061} / \textbf{297.02}} \\ \hline
\end{tabular}}
\end{table}

\section{Conclusion}\label{conclusions}
In this paper, we deal with energy time-series analysis and forecasting, with special emphasis to the Greek Energy Market. In particular, we focus to three different forecasting tasks, i.e., Electric Load Demand Forecasting, Personalized Energy Consumption Forecasting, and Renewable Energy Generation Forecasting. For each of the aforementioned tasks, recent state-of-the-art DL models are presented. 

\section*{Acknowledgments}
This work is co-ﬁnanced by the European Regional Development Fund of the European Union and Greek national funds through the Operational Program Competitiveness, Entrepreneurship and Innovation, under the call RESEARCH - CREATE - INNOVATE (project code: T2EDK-03048).

\bibliographystyle{unsrt}  
\bibliography{references}

\end{document}